\documentclass[runningheads]{llncs}
\usepackage{bbm}
\usepackage{graphicx}

\usepackage{tikz}
\usepackage{comment}
\usepackage{amsmath,amssymb} 
\usepackage{color}
\usepackage{tablefootnote}
\usepackage{booktabs}

\usepackage[accsupp]{axessibility}  

\makeatletter
\def\thanks#1{\protected@xdef\@thanks{\@thanks
        \protect\footnotetext{#1}}}
\makeatother

\begin{document}
\pagestyle{headings}
\mainmatter

\title{
MORE: Multi-Order RElation Mining \\ for Dense Captioning in 3D Scenes}


\titlerunning{MORE: Multi-Order RElation Mining for Dense Captioning in 3D Scenes}
%
\author{Yang Jiao\inst{1,2}$^\star$ \and
Shaoxiang Chen\inst{3}$^\star$\thanks{$^\star$ Equal contribution.} \and
Zequn Jie\inst{3} \and Jingjing Chen\inst{1,2}$^{\dagger}$, \\ Lin Ma\inst{3}$^{\dagger}$\thanks{$^\dagger$ Corresponding authors.} \and Yu-Gang Jiang\inst{1,2}}
\authorrunning{Jiao et al.}

\institute{Shanghai Key Lab of Intell. Info. Processing, School of CS, Fudan University 
\and
Shanghai Collaborative Innovation Center on Intelligent Visual Computing
\and
Meituan Inc.
}
\maketitle

\begin{abstract}
3D dense captioning is a recently-proposed novel task, where point clouds contain more geometric information than the 2D counterpart. However, it is also more challenging due to the higher complexity and wider variety of inter-object relations contained in point clouds. Existing methods only treat such relations as by-products of object feature learning in graphs without specifically encoding them, which leads to sub-optimal results. 
In this paper, aiming at improving 3D dense captioning via capturing and utilizing the complex relations in the 3D scene, we propose MORE, a Multi-Order RElation mining model, to support generating more descriptive and comprehensive captions. 
Technically, our MORE encodes object relations in a progressive manner since complex relations can be deduced from a limited number of basic ones.
We first devise a novel Spatial Layout Graph Convolution (SLGC), which semantically encodes several first-order relations as edges of a graph constructed over 3D object proposals. 
Next, from the resulting graph, we further extract multiple triplets which encapsulate basic first-order relations as the basic unit, and construct several Object-centric Triplet Attention Graphs (OTAG) to infer multi-order relations for every target object. 
The updated node features from OTAG are aggregated and fed into the caption decoder to provide abundant relational cues, so that captions including diverse relations with context objects can be generated. 
Extensive experiments on the Scan2Cap dataset prove the effectiveness of our proposed MORE and its components, and we also outperform the current state-of-the-art method. Our code is available at https://github.com/SxJyJay/MORE.

\end{abstract}

\section{Introduction}
Dense captioning, which aims at comprehending the visual scene through jointly localizing and describing multiple objects, has been extensively studied in the 2D computer vision community~\cite{Chen_2021_CVPR,chen2020learning,deng2021sketch,kim2019dense,wang2018bidirectional}. 
However, 2D data such as images and videos inherently lack the ability of accurately capturing the physical extent of objects and their locations in the scene. 
Recently, the 3D dense captioning task has been proposed by Chen et al.~\cite{chen2021scan2cap}, where pure point clouds are adopted as the visual representation to perform object localization and captioning on.
By connecting 3D scenes with natural language, 3D dense captioning has widespread application prospects in the field of human-machine interaction in augmented reality~\cite{kim2018revisiting,xiong2021augmented}, autonomous agents~\cite{savva2019habitat,xia2018gibson}, etc.

\begin{figure}[!t]
  \centering
  \includegraphics[width=0.9\linewidth]{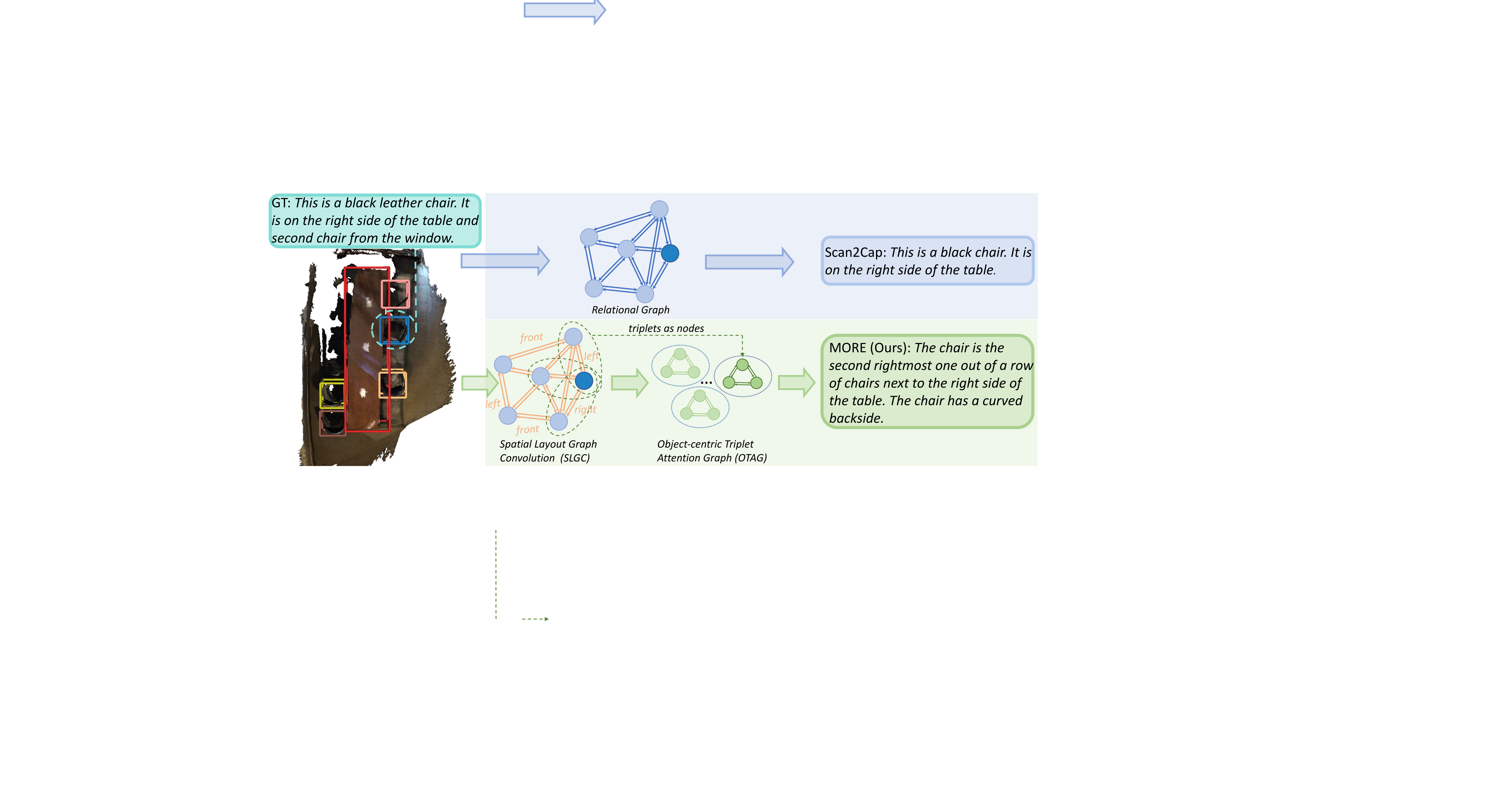}
  \caption{The comparison of our proposed Multi-Order RElation mining model (MORE) with the previous method (i.e., Scan2Cap~\cite{chen2021scan2cap}). The core components of our MORE and the Scan2Cap are distinguished with green and blue background, respectively. Scan2Cap treats the inter-object relations as by-products derived from node feature learning in the relational graph, thus the diverse spatial relations are under-explored. While in our MORE, we model such relations via a Spatial Layout Graph Convolution (SLGC) and Object-centric Triplet Attention Graphs (OTAG) to progressively encode more complex spatial relations. For simplicity, we only show the caption of one specific object (in the dashed circle) from different models. It is clear that our method can describe more complex relations.}
  \label{figure:fig1}
\end{figure}

The abundant geometric information contained in 3D point clouds can support describing object relations and the holistic scene (scene layouts) in a diversified manner, since the 3D point clouds are less limited by the occlusion and better capture object size and relative position.
For example, as shown in Fig.\ref{figure:fig1}, the circled chair can be described with diversiform relations like \textit{``on the right side''} and \textit{``second ... from ...''}. 
So in 3D dense captioning, mining the complex inter-object relations is of vital importance for generating comprehensive captions. 
But directly adapting relation modeling techniques in 2D images~\cite{pan2016jointly,pan2020x,chen2020zero,wang2020consensus,song2021spatial,ji2021step,jiao2021two,jiao2022suspected} to 3D point clouds can lead to poor performances as discussed in the previous work~\cite{chen2021scan2cap}. 
Scan2Cap, as the first 3D dense captioning work, develops a graph-based encoder to provide relation feature for the caption decoder and achieves promising results. 
However, it treats inter-object relations as by-products derived from node recognition in a graph, overlooking the gap between the visual world (represented by point clouds) and semantic concepts. 
Hence, complex relations, such as \textit{``between"}, \textit{``surrounded by"}, \textit{``rightmost"}, etc, can not be properly mapped into the semantic space, which leads to unitary descriptions being generated. 
As illustrated in Fig.\ref{figure:fig1}, Scan2Cap tends to describe the target object by capturing simple relations (\textit{``on the left"}).

To address the above problem, we propose MORE, a Multi-Order RElation mining model which can support generating more descriptive captions for objects in 3D point clouds. The core motivation behind MORE lies in the confidence that the multi-order relations can be deduced from a limited number of basic first-order relations. For example, given a scene where there are three objects in a row, dubbed \textit{A}, \textit{B}, and \textit{C}, and \textit{the relations between AB and BC are both ``on the left of''}, then the conclusion that \textit{C is the rightmost one} can be made. Hence, the main goal of our MORE is to first explicitly extract and encode basic first-order spatial relations among objects and then try to further infer multi-order relations for generating comprehensive captions.

Concretely, as shown in the Fig.\ref{figure:fig1} (green background part), MORE consists of two components: Spatial Layout Graph Convolution (SLGC) and Object-centric Triplet Attention Graphs (OTAG). 
First, to capture the concepts of basic first-order relation, SLGC adaptively introduces spatial semantics into the edges of an object graph. Afterward, inside the OTAG, triplets in the form of  $<$\textit{node$_1$,edge,node$_2$}$>$ are extracted from the previous graph as the basic descriptors of first-order inter-object relations, then several such object-centric triplet graphs are constructed where the triplets targeting the same object serve as nodes within the same triplet graph. On top of these graphs, the attention-based multi-order relation reasoning is performed within each of them. The advantage of using such triplets is that the basic relations are more explicitly preserved, and we can also flexibly extend a triplet to cover larger contexts.
Finally, our caption decoder receives the target object feature and performs node aggregation on OTAG as the context for sentence generation. As illustrated by our captioning results in Fig.\ref{figure:fig1}, the captions generated by MORE are more descriptive and comprehensive compared with the baseline method.

In summary, our contributions are threefold:
(1) We propose a Multi-Order RElation mining model (MORE) for 3D dense captioning, which can generate more descriptive and comprehensive captions for each object.
(2) Within the MORE, a Spatial Layout Convolution (SLGC) and Object-centric Triplet Attention Graphs (OTAG) are proposed and coupled together, where the former semantically encodes basic first-order spatial relations among objects in a 3D scene, and the latter infers the multi-order relations via attention-based graph reasoning.
(3) Extensive experimental results prove that our MORE achieves superior performances than existing 3D dense captioning methods on prevalent benchmarks.

\section{Related Work}
\textbf{Image and Video Captioning.}
Generating captions for 2D visual data (i.e., images and videos) has recently attracted significant research interest~\cite{Chen_2021_CVPR,chen2020learning,deng2021sketch,kim2019dense,wang2018bidirectional,yang2019auto,yao2018exploring,li2019know,wang2019role}. 
It is acknowledged that exploring the inter-object relationship benefits the caption generation, and such an idea has been widely investigated~\cite{yang2019auto,yao2018exploring,li2019know,wang2019role,yang2017dense,kim2019dense}. Yang et al.~\cite{yang2017dense} directly adopt the global image feature as context. To further introduce extra linguistic prior upon various object relations, scene graph detectors are utilized by some image captioning methods~\cite{yang2019auto,yao2018exploring} to parse the give image and assign textual tags to the relations. However, such detectors require expensive annotations to train. As an alternative, the part-of-speech tags are utilized as the prior to explicitly encode relations between objects in~\cite{kim2019dense}.
Although they are effective in 2D image and video caption generation,
the spatial structures are much more complicated in the 3D scene, hence they can not be directly transferred to 3D dense captioning task.

\noindent \textbf{3D Dense Captioning and Visual Grounding.}
Recently, investigating the 3D point cloud data and natural language has become a trending research topic~\cite{chen2020scanrefer,achlioptas2020referit3d,yuan2021instancerefer,huang2021text,feng2021free,chen2021scan2cap}. Among the pioneer works, Chen et al.~\cite{chen2020scanrefer} and Achloptas et al.~\cite{achlioptas2020referit3d} first proposed two datasets, referred to as ScanRefer and ReferIt3D, respectively, which both contain descriptions for real-world 3D objects in ScanNet~\cite{dai2017scannet}. On top of them, 3D visual grounding~\cite{chen2020scanrefer,achlioptas2020referit3d,yuan2021instancerefer,huang2021text,feng2021free} and 3D dense captioning~\cite{chen2021scan2cap}, as two dual tasks, are concurrently investigated, where the former focuses on localizing 3D objects described by natural language queries and the latter aims at generating descriptions for 3D objects in RGB-D scans. Exploring object relations is essential for both tasks, and many relevant attempts have been made in recent works~\cite{yuan2021instancerefer,huang2021text,feng2021free,he2021transrefer3d,zhao20213dvg,yang2021sat,chen2021scan2cap}. 

In the 3D visual grounding task, earlier works, namely TGNN~\cite{huang2021text} and InstanceRefer~\cite{yuan2021instancerefer}, construct a directed instance graph with instance features as vertices and relative instance coordinates as edges. Later, in order to capture the object-object and object-scene co-occurrence, FFD~\cite{feng2021free} develops a multi-level proposal relation graph module with a geometric structure feature of each bounding box encoded in each graph node. 
Aiming at a unified intra- and inter-modalities modeling scheme, Transfer3D~\cite{he2021transrefer3d}, 3DVG~\cite{zhao20213dvg}, and SAT~\cite{yang2021sat} adopt a standard Transformer architecture~\cite{vaswani2017attention} for promoting inter-object relations. However, these methods overlooked the importance of encoding the multi-order visual object relations, which might be because that the key to improve grounding performance is learning the explicit correspondence between vision and language.
And existing methods still struggle at establishing such correspondence due to the lack of semantics in the point clouds data~\cite{yang2021sat}.

In the 3D dense captioning task, Scan2Cap~\cite{chen2021scan2cap} first proposes a relational graph implemented with a static version of EdgeConv~\cite{wang2019dynamic} to enhance object relation representation. However, the inter-object relations are only treated as by-products of graph node recognition, thus leads to sub-optimal results. Theoretically, 3D visual relation detectors~\cite{wald2020learning,armeni20193d,wu2021scenegraphfusion} can mitigate such the problem, however, the current results delivered by them are unsatisfactory when faced with highly unrestricted scene compositions~\cite{zhang2018grounding,milewski2020scene}. 
Therefore, in this paper, we aim to develop a relation mining method which can properly encode both the basic first-order and complex multi-order relations contained in the 3D scene, so as to benefit more comprehensive caption generation.

\begin{figure}[!t]
  \centering
  \includegraphics[width=\linewidth]{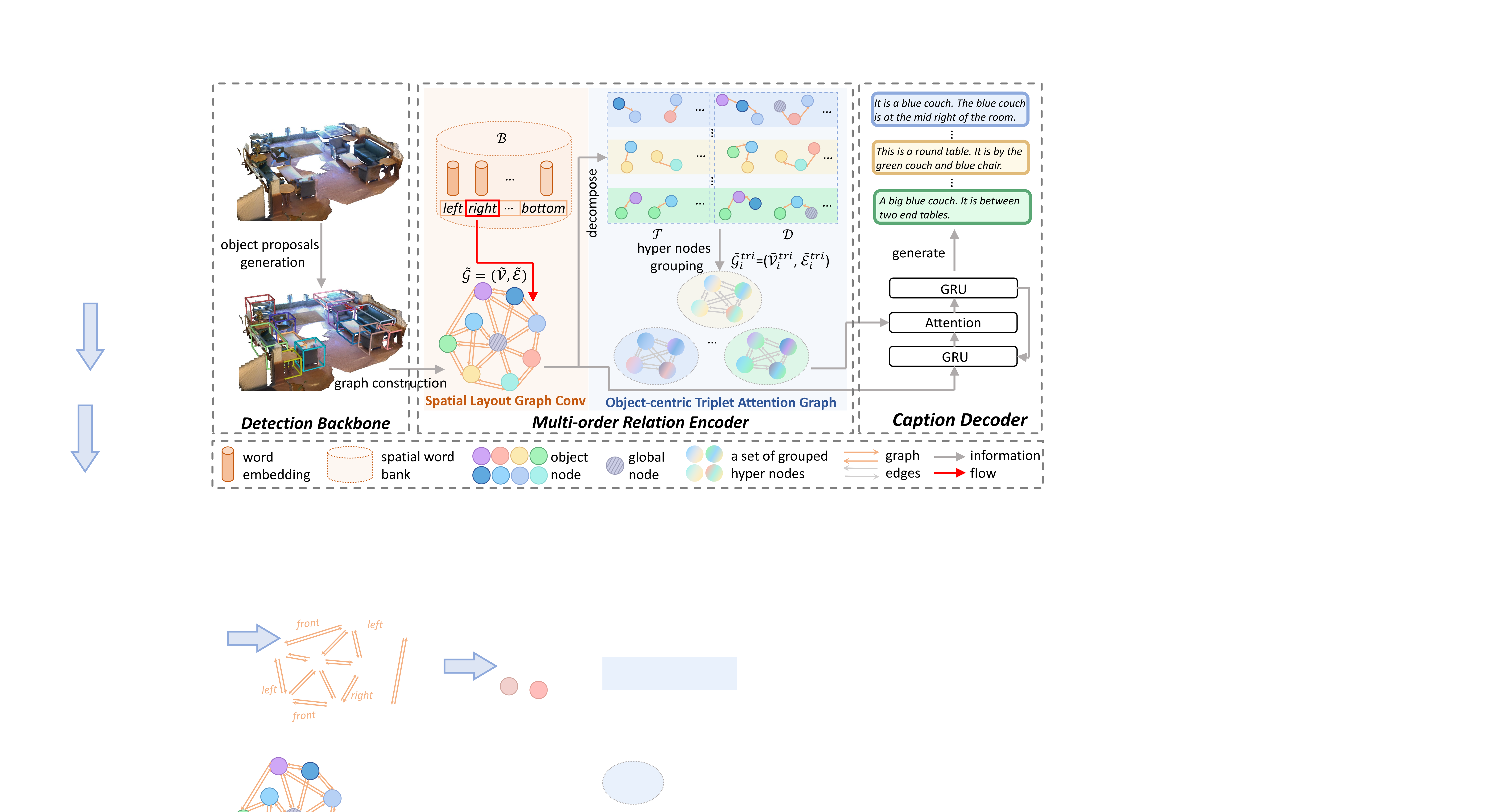}
  \caption{The overall framework of our proposed method, which consists of three parts: the detection backbone, the multi-order relation encoder, and the caption decoder. Given a 3D scene represented by point clouds, the detection backbone extracts a set of object proposals. Then, based on the objects, first-order and multi-order spatial relations are progressively encoded through a novel Spatial Layout Graph Convolution (SLGC) and several Object-centric Triplet Attention Graphs (OTAG), respectively. Finally, the OTAG's output, which encapsulates rich spatial relational cues, are served as the context to aid comprehensive caption generation. To keep the figure concise, we omit part of the object nodes and their corresponding captions, as well as the attention calculation of each triplet graph.}
  \label{figure:fig2}
\end{figure}

\section{Multi-order Relation Mining Network}
The overall framework of our Multi-Order RElation mining (MORE) network consists of three main components: an object detection backbone, a multi-order relation encoder, and a caption decoder.
As shown in Fig.\ref{figure:fig2}, given a point cloud as input, the detection backbone extracts several 3D object proposals with bounding boxes (Sec.\ref{subsec:3.1}).
Then, as the core component of our framework, the multi-order relation encoder first takes object proposals as input and explicitly encodes the first-order spatial relations via our proposed Spatial Layout Graph Convolution (SLGC).
During this process, SLGC maintains a bank of spatial words and dynamically selects word embeddings from it as graph edges, so as to narrow the gap between point cloud and relational concepts. 
The resulting graph is decomposed into triplets and recomposed as several Triplet Object-centric Attention Graphs (OTAG) for multi-order spatial relation reasoning (Sec.\ref{subsec:3.2}).
Finally, the decoder takes updated node features from OTAG as inputs to incorporate contextual cues with an attention module and generates language descriptions for the corresponding objects (Sec.\ref{subsec:3.3}).
\subsection{Detection Backbone}
\label{subsec:3.1}
Given a 3D scene input represented by a point cloud $\mathcal{P}=\{(p_i,f_i)\}_{i=1}^{N_P}$ ($N_P$ is the number of points), where $p_i \in \mathbb{R}^3$ and $f_i \in \mathbb{R}^{3}$ are the coordinates (\textit{x-y-z}) and the color (\textit{r-g-b}) of the $i$-th point, respectively, we adopt the PointNet++~\cite{qi2017pointnet++} backbone and the voting module in VoteNet~\cite{qi2019deep} to extract a set of object proposals.
We denote these object proposals as $\mathcal{O}=\{(x_i,c_i)\}_{i=1}^{N_O}$ ($N_O$ is the number of valid object proposals), where $x_i \in \mathbb{R}^{128}$ is the $i$-th object proposal's visual feature and $c_i=(b^x,b^y,b^z,b^h,b^w,b^l)$ is the location indicator of the corresponding bounding box. $b^x,b^y,b^z$ are the coordinates of the box center and $b^h,b^w,b^l$ are height, width, length of the box, respectively.
\subsection{Multi-order Relation Encoding}
\label{subsec:3.2}
The key to generating accurate and diverse captions for 3D real world is to capture or infer the complicated spatial relations, however, this is overlooked by the previous method~\cite{chen2021scan2cap}.
As the core component of our method, the multi-order relation encoder takes object proposals $\mathcal{O}$ as input, and is responsible for explicitly encoding the basic first-order spatial relations by Spatial Layout Graph Convolution (SLGC) and further inferring possible multi-order relations via Object-centric Triplet Attention Graphs (OTAG).
In the rest of this section, we will elaborate on each of them.

\noindent \textbf{Spatial Layout Graph Convolution.}
Following~\cite{chen2021scan2cap}, we wish to connect each object with its $K$ nearest neighboring instances, and thereby construct an object graph to represent the spatial layout of the whole scene, denoted as $\mathcal{G}=(\mathcal{V},\mathcal{E})$.
Note that in addition to  the object proposals, we also add a global node into $\mathcal{V}$ to model the interaction between each object and the whole scene.
Formally, $\mathcal{V}=\{v_i\}_{i=1}^{N_O+1}$, where $v_j=x_j$ ($j=1,\dots,N_O$) and $v_{N_O+1}=g$. $g$ is the feature of global node calculated by averaging all object proposals' features. 

As for the edges, conventional methods~\cite{chen2021scan2cap,feng2021free} treat the graph edges $\mathcal{E}$ as the by-products derived from node representation learning, thus the edges do not have clear semantics.
So we construct our spatial layout graph by delicately encoding basic first-order spatial relational concepts that are transferred from the linguistic embedding space. 

First, we maintain a spatial word bank $\mathcal{B}$ to provide semantic concepts of basic first-order spatial relations:
\begin{equation}
\label{eq1}
    \mathcal{B} = \{``left",``right",``front",``behind",``besides",``top",``bottom"\}.
\end{equation}
For illustration purpose, we list the spatial words in $\mathcal{B}$, while we use their corresponding GloVE~\cite{pennington2014glove} word embeddings in our model. These relations can be divided into horizontal and vertical subsets, $\mathcal{B}_h$ and $\mathcal{B}_v$, respectively\footnote{``top'' and ``bottom'' are the vertical relations we focus on.}.

Next, we need to select a corresponding item from the word bank as the edge to describe the spatial relationship of the two connected nodes, however, this is nontrivial in the 3D world, as the varying viewpoints may bring ambiguity to the horizontal spatial relations.
To this end, we select the horizontal spatial word in an adaptive manner, and the less ambiguous vertical spatial word based on rules.
To compute the horizontal spatial relation between objects $i$ and $j$, we first calculate the relative coordinates $c_{j,i}$ and object feature $v_{j,i}$, and combine them to predict a distribution $\alpha_{j,i} \in \mathbb{R}^{N_h}$ of the horizontal relational words:
\begin{equation}
\label{eq2}
    \begin{split}
        & v_{j,i} = W_v[v_i;v_j-v_i], \quad c_{j,i} = W_c[b^x_j-b^x_i;b^y_j-b^y_i], \\
        & \alpha_{j,i} = \mathrm{softmax}(W_{\alpha}\mathrm{tanh}(v_{j,i}+c_{j,i})),
    \end{split}
\end{equation}
where $[;]$ represents the concatenation operation, and $W_v \in \mathbb{R}^{256 \times 128}$, $W_c \in \mathbb{R}^{2 \times 128}$, and $W_{\alpha} \in \mathbb{R}^{128 \times N_h}$ are weight matrices. Afterward, the horizontal spatial relational feature $r^{h} \in \mathbb{R}^{300}$ can be calculated by a weighted combination of the corresponding items of the spatial word bank:
\begin{equation}
\label{eq3}
    r^{h} = \alpha_{j,i} \cdot \mathcal{B}_h,
\end{equation}
where $\cdot$ is the dot product operation. 
In this way, 

As for the vertical spatial relations, namely ``top'' and ``bottom'', we design a metric to directly infer vertical relation between a given pair of objects $i$ and $j$. We first generate the box corner coordinates $b^c_i, b^c_j \in \mathbb{R}^{8}$ according to their initial box coordinates $c_i$ and $c_j$. Then, we calculate the relative vertical distance between each pair of corner points between objects $i$ and $j$, and obtain the pairwise distances $d_{j,i}^v \in \mathbb{R}^{64}$. 
The vertical spatial relations can then be inferred by taking the sign of the maximum or the minimum element of $d_{j,i}^v$. And we can formulate the process of obtaining vertical spatial relational feature $r^v$ as:
\begin{equation}
\label{eq4}
    r^v = \mathbb{I}(\mathrm{min}(d^v_{j,i})>0)\times \mathcal{B}_v^{top}+\mathbb{I}(\mathrm{max}(d^v_{j,i})<0) \times \mathcal{B}_v^{bottom}
\end{equation}
where $\mathbb{I}(\cdot)$ is the indicator function, which equals 1 when the condition inside the bracket is satisfied, otherwise 0, and $\mathcal{B}_v^{top}$ and $\mathcal{B}_v^{bottom}$ are the corresponding items in the word bank, respectively. Finally, the first-order spatial relation between object $i$ and $j$, also known as the directed graph edge $e_{j,i}$, can be obtained through combining two relational features $r^{h}$ and $r^{v}$:
\begin{equation}
\label{eq5}
    e_{j,i} = W_e[r^v;r^h],
\end{equation}
where $W_e \in \mathbb{R}^{600 \times 128}$ is the projection matrix.
As for the global node $g$, we manually assign its center coordinates as the location of the scene's center for the horizontal relational feature calculation. 
And the vertical relational feature is an all-zero vector due to that no bounding box is available for the whole scene.

At this point, we have constructed the initial spatial layout graph $\mathcal{G}$ and the basic first-order spatial relations between objects have been incorporated into the graph edges via the semantically rich word embeddings.
Then, we perform message passing upon the spatial layout graph to enhance the node features by letting them aggregate information from neighboring nodes:
\begin{equation}
\label{eq6}
\begin{split}
     &\beta_{j,i} = {(W_1v_i)^{\mathrm{T}}}(W_2v_j+W_3e_{j,i}), \quad \hat{\beta}_{j,i}=\frac{\mathrm{exp}(\beta_{j,i})}{\sum_{k \in \mathcal{N}(i)}\mathrm{exp}(\beta_{k,i})}, \\
    & v_i' = W_4v_i + \sum_{j \in \mathcal{N}(i)}\hat{\beta}_{j,i}\odot(W_5v_j+W_6e_{j,i}),
\end{split}
\end{equation}
where the $\odot$ represents the element-wise multiplication with broadcast operation.
Note that when computing aggregation weights, we also take the semantic edges into consideration. 
The above Spatial Layout Graph Convolution (SLGC) can be stacked multiple times and we denote the final resulting graph as $\widetilde{\mathcal{G}}=(\widetilde{\mathcal{V}},\widetilde{\mathcal{E}})$, whose node and edge features can be denoted as $\widetilde{\mathcal{V}}=\{\widetilde{v}_i\}_{i=1}^{N_O+1}$ and $\widetilde{\mathcal{E}}=\{\widetilde{e}_{i,j}\}_{i,j=1}^{N_O+1}$.

\noindent \textbf{Object-centric Triplet Attention Graph.}
Theoretically, by conducting multiple rounds of message passing with stacked SLGC, the model can become aware of the multi-order relations, however, we find in our experiments (Table.\ref{table:comprehensive}) that this can not achieve the expected effects since the captioning performance does not improve with more layers.
We postulate such a phenomenon might be caused by the over-smoothing problem~\cite{chen2020measuring,zhou2020towards}, and the experimental results (will be analyzed in detail) in Table.\ref{table:comprehensive} support our assumption to some extent. 
Thus we decide to not solely rely on the SLGC, but further specifically design an object-centric graph that preserves more information of the target objects and the relations.

Concretely, based on the first-order spatial relation aware graph $\widetilde{\mathcal{G}}$ that we obtained from SLGC, we directly extract a set of $<$\textit{node$_1$,edge,node$_2$}$>$ triplets from it, and the \textit{node}$_2$ is the target object that we wish to generate caption for.
Formally, these triplets can be represented as $\mathcal{T}=\{<\widetilde{v}_j,\widetilde{e}_{j,i},\widetilde{v}_i>\}_{j \in \mathcal{N}(i), i=1:N_O}$.
Note that we do not extract triplets targeting the global node, since currently we are focusing on generating captions for the objects but not the whole scene. 
Such triplets are a combined representation of a target object and its surrounding contexts, as well as their relations, and the target object is emphasized in each triplet graph so that its information will not be overwhelmed in the following operations.
To enlarge the context, we can further extend these triplets via constructing a new set of quintuplets
which can be represented as
$\mathcal{Q}=\{(\widetilde{v}_k,\widetilde{e}_{k,j},\widetilde{v}_j,\widetilde{e}_{j,i},\widetilde{v}_i)\}_{k \in \mathcal{N}(j), j \in \mathcal{N}(i), i=1:N_O}$. In practice, we find these extended quintuplets to be helpful.

We then take the triplets/quintuplet ($\mathcal{T}$/$\mathcal{Q}$) as hyper-nodes, and construct object-centric graphs upon them. The triplets/quintuplets targeting the same object node are regarded as related hyper-nodes and we will further model their relations as a object-centric triplet attention graph.
We denote the recomposed triplet graph that is centered at object $i$ as  $\mathcal{G}^{tri}_{i}=(\mathcal{V}^{tri}_{i},\mathcal{E}^{tri}_{i})$ ($i=1,\dots,N_O$).
Since the hyper-nodes from $\mathcal{T}$ and $\mathcal{Q}$ have different formats, we first unify their channel dimensions by mapping them into the same lower-dimensional space:
\begin{equation}
\label{eq7}
    v_{j,i} = W_{t}[\widetilde{v_j};\widetilde{e}_{j,i};\widetilde{v_i}], \quad v_{k,j,i} = W_{q}[\widetilde{v_k};\widetilde{e}_{k,j};\widetilde{v_j};\widetilde{e}_{j,i};\widetilde{v_i}],
\end{equation}
where $W_{t} \in \mathbb{R}^{384 \times 128}$ and $W_{q} \in \mathbb{R}^{640 \times 128}$ are projection matrices.
Then, the hyper-nodes can be represented as $\mathcal{V}^{tri}_{i}=\{v_{j,i}\}_{j \in \mathcal{N}(i)}\cup \{v_{k,j,i}\}_{k \in \mathcal{N}(j),j \in \mathcal{N}(i)}$.
As for the edges $\mathcal{E}^{tri}_{i}$, we connect each pair of nodes in $\mathcal{V}^{tri}_i$ by computing the edge weights via attention:
\begin{equation}
    e_{\{j,i\},\{k,i\}}= \frac{\exp(\sigma(W_7 v_{j,i}) \sigma(W_8 v_{k,i}))}{\mathcal{Z}_{j,i}},
\end{equation}
where $\{j,i\}$ and $\{k,i\}$ represents two hyper-nodes of triplets (or quintuplets), $W_7, W_8$ are projection matrices, and $\sigma$ is an activation function (e.g., ReLU). We denote the normalization denominator as $\mathcal{Z}_{j,i}$ to avoid clutter.
We then aggregate neighbor information through
\begin{equation}
    v'_{j,i} = \sum_{k \in \mathcal{N}(i)} e_{\{j,i\},\{k,i\}} W_9 v_{k,i}.
\end{equation}
Here we only include triplet-based hyper-nodes for clarity, and quintuplets can be similarly incorporated. 
By performing the above message passing among hyper-nodes, the model can infer multi-order relations while significantly preserving the target object's information.
The final object-centric triplet graph after pair-wise node attention calculation (i.e., Eq.(\ref{eq8}) and Eq.(\ref{eq9})) is denoted as $\widetilde{\mathcal{G}}^{tri}_{i}=(\widetilde{\mathcal{V}}^{tri}_i,\widetilde{\mathcal{E}}^{tri}_{i})$ ($i=1,\dots,N_O$).
\subsection{Caption Generation}
\label{subsec:3.3}
It is a common practice to make the caption decoder responsible for incorporating contextual cues generated by the encoder~\cite{anderson2018bottom,chen2021scan2cap}.
We design our decoder based on a two-layer GRU network, and equip it with an attention module in between the two layers to aggregate OTAG's object-centric nodes that are aware of multi-order spatial relations.

In particular, the caption is iteratively generated in a word-by-word manner. 
At the $t$-th time step, the first GRU takes the concatenation of the GloVE embedding of the word $w^{(t-1)}$, the previous hidden state $h^{(t-1)}_2$ of the second GRU, as well as the initial visual feature $x_i$ of the target object proposal as inputs, and
update the its hidden state $h^{(t)}_1$ as:
\begin{equation}
\label{eq8}
    h^{(t)}_1 = \mathrm{GRU}_1([w^{(t-1)};h^{(t-1)}_2;x_i]; h^{(t-1)}_1),
\end{equation}
Next, depending on the updated hidden state $h^{(t)}_1$ of the first GRU, 
we adaptively compute aggregation weights for $\widetilde{\mathcal{G}}^{tri}_i$'s node features, which contains sptatial relation cues between the target object and its neighbors, and these node features are then aggregated by a weighted sum:
\begin{equation}
\label{eq9}
    \begin{split}
        \gamma_i = W_{\gamma}(\mathrm{tanh}(W_{10}&h^{(t)}_1 +W_{11}\widetilde{v}^{tri}_i)), \quad \hat{\gamma}_i = \frac{\mathrm{exp}(\gamma_i)}{\sum_{j \in \mathcal{N}(i)}\mathrm{exp}(\gamma_j)}, \\
        & \hat{v}^{(t)} = \sum_{i=1}^{N_O}\hat{\gamma}_i\odot\widetilde{v}^{tri}_i,
    \end{split}
\end{equation}
where $\hat{v}^{(t)}$ is the resulting contextual feature vector with spatial relations embedded in it. Afterward, the contextual vector $\hat{v}^{(t)}$, together with the hidden state $h^{(t)}_1$ of the first GRU, are fed into the second GRU to obtain its updated hidden state $h^{(t)}_2$:
\begin{equation}
\label{eq10}
    h^{(t)}_2 = \mathrm{GRU}_2([\hat{v}^{(t)};h^{(t)}_1]; h^{(t-1)}_2),
\end{equation}
Finally, $h^{(t)}_2$ is leveraged to predict the current word $w^{(t)}$ through a linear classifier. 
The details about the captioning loss and model training process can be found in the supplementary materials.

\section{Experiments}
\subsection{Experimental Setup}
\textbf{Dataset.}
Following previous work~\cite{chen2021scan2cap,chen2021d3net}, we use the ScanRefer~\cite{chen2020scanrefer} dataset, which consists of 51,583 descriptions for 11,046 objects in 800 ScanNet~\cite{dai2017scannet} scenes. 
The descriptions include the appearance of the objects (e.g. “this is a black tv”), and the spatial relations between the target object and surrounding objects (e.g. “the cabinet is next to the desk”). 
The dataset is split into train/val sets with 36,665 and 9,508 samples respectively following the  ScanRefer~\cite{chen2020scanrefer} benchmark. Scenes in the train and val sets are disjoint with each other. Since the test set has not been officially released, all experimental results and analysis in the following sections are conducted on top of the val set. 

\noindent \textbf{Evaluation Metrics.}
To jointly evaluate the quality of detected bounding boxes and generated captions, a combined metric $m@k\mathrm{IoU}=\frac{1}{N}\sum_{i=1}^{N}m_{i}u_{i}$ defined in~\cite{chen2021scan2cap} is adopted, where $u_{i}$ is set to 1 if the IoU score for the $i^{th}$ box is greater than $k$, otherwise 0, and $m$ can be one of the caption metrics, such as CiDEr~\cite{vedantam2015cider}, BLEU-4~\cite{papineni2002bleu}, METEOR~\cite{banerjee2005meteor}, and ROUGE~\cite{lin2004rouge}, abbreviated as C, B-4, M, R in the following part, respectively. Meanwhile, mean average precision (mAP) at specified IoU threshold is utilized to evaluate the object detection performance.

\noindent \textbf{Implementation Details.} 
In our experiment, we randomly sample 40,000 points from every ScanNet scenes. The maximum number of object proposals $K$ is set as 256. Unless otherwise specified, we utilize the color (\textit{r-g-b}) of each point as the input visual feature to conduct experiments.
The number of stacked SLGC layers are set as 1 and 2 when the point color and multi-view feature are adopted, respectively.
We train our model with the Adam optimizer~\cite{kingma2014adam}, and set the learning rate to $1e^{-4}$, weight decay to $1e^{-5}$, and batch size to 12.
Following~\cite{chen2021scan2cap}, we adopt the same data augmentation strategy and truncate the descriptions longer than 30.
\subsection{Comparison with State-of-the-art}
We compare our method with the state-of-the-art approach Scan2Cap~\cite{chen2021scan2cap} on the ScanRefer~\cite{chen2020scanrefer} benchmark as shown in Table.\ref{table:SOTA}, where the VoteNet~\cite{qi2019deep} is adopted as the detector and the entire network is trained end-to-end. 
Note that the results that we reproduced with the officially released code of Scan2Cap~\cite{chen2021scan2cap} have discrepancy with the results reported in the paper, and we report our reproduced results in Table.\ref{table:SOTA}, denoted as ``Scan2Cap*''. The methods in the first three rows are simple baselines provided in the previous work~\cite{chen2021scan2cap}.
The results demonstrate that our method achieves consistent improvements across most of the evaluation metrics, especially on CiDEr. 
When using \textit{``r-g-b"} color and multiview features as additional point features, compared with Scan2Cap, our method obtains 5.16\% and 6.09\% improvements on C@0.25IoU, as well as 3.78\% and 1.86\% improvements on C@0.5IoU in respectively.
We emphasize the CIDEr metric because compared with other evaluation metrics, CiDEr shows higher agreement with consensus as assessed by humans~\cite{vedantam2015cider}.
Overall, the significant improvements on C@0.25IoU and C@0.5IoU can prove the superiority of our method.

Meanwhile, considering that the overall performances of the dense captioning can be affected by the detector, we adopt ground-truth bounding boxes as input to specifically compare the captioning performance as shown in Table.\ref{table:SOTA_GT}. Methods in the first three lines are simple baselines provided in~\cite{chen2021scan2cap}.
Since now the performance is no longer affected by the detection results, we omit the mAP and captioning results with the 0.25 IoU threshold. 
We observe from the table that our method surpasses the baseline with a large margin across all evaluation metrics.
Besides, the improvements over Scan2Cap are more evident than in Table.\ref{table:SOTA} where VoteNet is used as the detector. This indicates that when the context objects are more reliable, our method can benefit more and perform relation encoding with a higher quality, thereby promoting the captioning performances.
\subsection{Comprehensive Analysis}
\textbf{Ablation Studies.} We conduct ablation studies to verify the effectiveness of our proposed SLGC and OTAG designs, and the results are shown in Table.\ref{table:ablation}. 
In the baseline method (the first row), we remove OTAG and the edge feature $e_{j,i}$ in Eq.\eqref{eq6} when performing message passing and keep other settings of SLGC the same. 
As shown in the second row, including SLGC improves C@0.5IoU by 1.79\% comparing to the baseline, and other metrics are also improved.
\begin{table*}[!t]
\centering
\caption{Comparison with state-of-the-art methods on the ScanRefer dataset with VoteNet~\cite{qi2019deep} as the detector of all methods. ``Scan2Cap*'' represents the results we reproduced with the officially released code. We use subscript ``rgb" and ``mul" to denote using \textit{``r-g-b"} color and multi-view feature as additional point features.}
\label{table:SOTA}
\scalebox{0.7}{
\begin{tabular}{c|cccc|cccc|c}
\toprule
Method    & C@0.25IoU & B-4@0.25IoU & M@0.25IoU & R@0.25IoU & C@0.5IoU       & B-4@0.5IoU     & M@0.5IoU       & R@0.5IoU       & mAP@0.5 \\ \hline
2D-3D Proj.~\cite{chen2021scan2cap}  & 18.29     & 10.27       & 16.67     & 33.63     & 8.31          & 2.31          & 12.54          & 25.93          & 10.50   \\
3D-2D Proj.~\cite{chen2021scan2cap}  & 19.73     & 17.86       & 19.83     & 40.68     & 11.47          & 8.56          & 15.73          & 31.65          & 31.83   \\
VoteNetRetr~\cite{qi2019deep}  & 15.12     & 18.09       & 19.93     & 38.99     & 10.18          & 13.38          & 17.14          & 33.22          & 31.83   \\ \hline \hline
Scan2Cap$_{mul}$~\cite{chen2021scan2cap}  & 56.82     & 34.18       & 26.29     & 55.27     & 39.08          & \textit{\textbf{23.32}}          & \textit{\textbf{21.97}}          & \textit{\textbf{44.78}}          & 32.21   \\
Scan2Cap*$_{mul}$~\cite{chen2021scan2cap} & 53.88         & 32.71           & 25.64         & 53.87         & 38.11          & 22.63          & 21.60          & 44.06              & 31.47       \\
MORE$_{mul}$      & \textit{\textbf{62.91}}         & \textit{\textbf{36.25}}           & \textit{\textbf{26.75}}         & \textit{\textbf{56.33}}         & \textit{\textbf{40.94}} & 22.93 & 21.66 & 44.42 & \textit{\textbf{33.75}}  \\ \hline \hline     
Scan2Cap$_{rgb}$~\cite{chen2021scan2cap}\tablefootnote{Results of this setting can be found in the supplementary materials of~\cite{chen2021scan2cap}.}  & 53.73     & 34.25       & 26.14     & 54.95     & 35.20          & 22.36          & 21.44          & 43.57          & 29.13   \\
Scan2Cap*$_{rgb}$~\cite{chen2021scan2cap} & 51.05         & 32.99           & 25.59         & 53.82         & 35.11          & 22.26          & 21.44          & 43.70              & 28.86       \\
MORE$_{rgb}$      & \textbf{58.89}         & \textbf{35.41}           & \textbf{26.36}         & \textbf{55.41}         & \textbf{38.98} & \textbf{23.01} & \textbf{21.65} & \textbf{44.33} & \textbf{31.93}\\ \bottomrule
\end{tabular}}
\end{table*}

\begin{table}[!t]
\centering
\caption{Comparison with state-of-the-art methods on the ScanRefer dataset using ground-truth bounding boxes for all methods. Since the experimental results using ground-truth bounding boxes with \textit{``r-g-b''} as additional point features are not reported by Scan2Cap, we only report our reproduced results with their officially released code.}
\label{table:SOTA_GT}
\scalebox{0.72}{
\begin{tabular}{c|cccc}
\toprule
          & C@0.5IoU       & B-4@0.5IoU     & M@0.5IoU       & R@0.5IoU       \\ \hline
OracleRetr2D~\cite{chen2021scan2cap} & 20.51          & 20.17          & 23.76          & 50.98          \\
Oracle2Cap2D~\cite{chen2021scan2cap} & 58.44          & 37.05          & 28.59          & 61.35          \\
OracleRetr3D~\cite{chen2021scan2cap} & 33.03          & 23.36          & 25.80          & 52.99          \\ \hline \hline
Scan2Cap$_{mul}$~\cite{chen2021scan2cap} & 67.95          & 41.49          & 29.23          & 63.66          \\
Scan2Cap*$_{mul}$~\cite{chen2021scan2cap} & 65.51          & 39.62          & 29.23          & 62.87          \\
MORE$_{mul}$      & \textbf{70.39} & \textbf{42.34} & \textbf{29.55} & \textbf{64.31} \\ \hline \hline
Scan2Cap*$_{rgb}$~\cite{chen2021scan2cap} & 64.19          & 38.90          & 28.96          & 62.38          \\
MORE$_{rgb}$      & \textbf{67.15} & \textbf{43.52} & \textbf{29.55} & \textbf{65.09}  \\ \bottomrule
\end{tabular}}
\end{table}
This demonstrates that explicitly encoding the basic first-order spatial relational concepts when updating node features is helpful to improving the captioning accuracy.
In the third row, OTAG alone brings 4.28\% C@0.5IoU improvement over the baseline, which indicates that these recomposed object-centric graphs could effectively enhance our model's ability to capture multi-order relations, thus significantly boosting the performance. 
Combining the SLGC and OTAG, as shown in the last row, can achieve the highest performance, especially for the C@0.5IoU (38.98\%), which outperforms the baseline with a large margin of 5.31\%.
These experiments demonstrate that the key components of our model (SLGC and OTAG) are both beneficial to generating high quality dense captions, and that our multi-order encoding of the inter-object relations is effective. 

\begin{table}[!t]
\centering
\caption{Ablation studies of the individual components, including the SLGC for first-order relation encoding, and the OTAG for multi-order relation modeling.}
\label{table:ablation}
\scalebox{0.72}{
\begin{tabular}{cc|ccccc}
\toprule
SLGC &OTAG & C@0.5IoU       & B-4@0.5IoU     & M@0.5IoU       & R@0.5IoU       & mAP@0.5        \\ \hline
     &    & 33.67          & 20.92         & 20.87          & 42.60         & 29.14             \\
\checkmark     &    & 35.46          & 21.31          & 21.01          & 43.10          & 29.18              \\
     & \checkmark   & 37.95              & 21.55              & 21.17              & 43.49              & 29.53              \\
\checkmark     & \checkmark    & \textbf{38.98} & \textbf{23.01} & \textbf{21.65} & \textbf{44.33} & \textbf{31.93} \\ \bottomrule
\end{tabular}}
\end{table}

\begin{table}[t]
\centering
\caption{Comparison of different configurations of message passing for multi-order relation modeling in the graph structure.}
\label{table:comprehensive}
\scalebox{0.72}{
\begin{tabular}{c|ccccc|c}
\toprule
           & C@0.5IoU       & B-4@0.5IoU     & M@0.5IoU       & R@0.5IoU       & mAP@0.5    & MADGap \\ \hline
SLGC$\times$1     & 35.46          & 21.31          & 21.01          & 43.10          & 29.18       & 17.46\\
SLGC$\times$2     & 38.06          & 22.26          & 21.35          & 43.64          & 29.98       & 14.49\\
SLGC$\times$3     & 38.27          & 21.99          & 21.16          & 43.43          & 30.43       & 13.17\\
SLGC$\times$4     & 36.58          & 21.46          & 21.05          & 42.93          & 29.94       & 12.28\\
SLGC$\times$1+OTAG & \textbf{38.98} & \textbf{23.01} & \textbf{21.65} & \textbf{44.33} & \textbf{31.93} & \textbf{21.70}\\
\bottomrule
\end{tabular}}
\end{table}

\noindent \textbf{Why not stacking SLGC for multi-order relation modeling?}
Theoretically, multi-order relations can be modeled by message passing on a graph for several rounds so that one node can reach out to further neighbors, but as our studies in Table.\ref{table:comprehensive} demonstrate, such an approach cannot attain ideal outcomes.
In the first four rows, we stacked SLGC of 1, 2, 3, and 4 layers, respectively, and then feed the output nodes directly to the the caption generation decoder as in Scan2Cap~\cite{chen2021scan2cap}.
We can see that although stacking two layers of SLGC can bring clear improvements over using only one SLGC layer
However, adding more layers of SLGC hurts the model's performance. When the number of the SLGC layers increase to 3 and 4, the model's performance gradually degrades.
Since stacking SLGC only changes the node features we fed to the caption decoder, we conjecture that the performance degradation might be due to the over-smoothing issue as described  in~\cite{chen2020measuring,zhou2020towards}.
We will further analyze this issue and provide evidence in the subsequent paragraph. 
Next, instead of stacking multiple layers of SLGC, we stack our OTAG on top of 1-layer SLGC.
The results are shown in the last row of Table.\ref{table:comprehensive}, and by comparing it with the SLGC-only results, we can observe that SLGC$\times$1+OTAG outperforms the best results of stacking SLGC (SLGC$\times$2). 
This proves that constructing OTAG is a more suitable approach of modeling multi-order relations under our scenario. 

\noindent \textbf{Why is OTAG more suitable?}
As shown in Eq.\eqref{eq8},\eqref{eq9},\eqref{eq10}, the caption generation is conditioned on the target object feature $x_i$ and its context feature $\hat{v}^{(t)}$, and the context feature is adaptively computed based on the decoding state and the encoder's graph nodes. 
Hence, obtaining distinctive node feature representation $\hat{v}^{(t)}$ from the multi-order relation encoder is the key to generate diverse captions. 
However, if the multi-order relation encoder is simply composed of multiple graph convolution layers, the node representation might suffer from the over-smoothing issue~\cite{chen2020measuring,zhou2020towards} (i.e., features of the graph nodes from different classes would become indistinguishable when stacking multiple graph layers~\cite{chen2020measuring}), which hurts obtaining distinctive context features. Hence we conduct experiments to verify our OTAG is a more suitable graph layer that can learn more distinguishable node features to benefit the caption decoder. 

\begin{figure}[t]
    \centering
    \begin{minipage}[t]{0.45\linewidth}
        \centering
        \includegraphics[width=\linewidth]{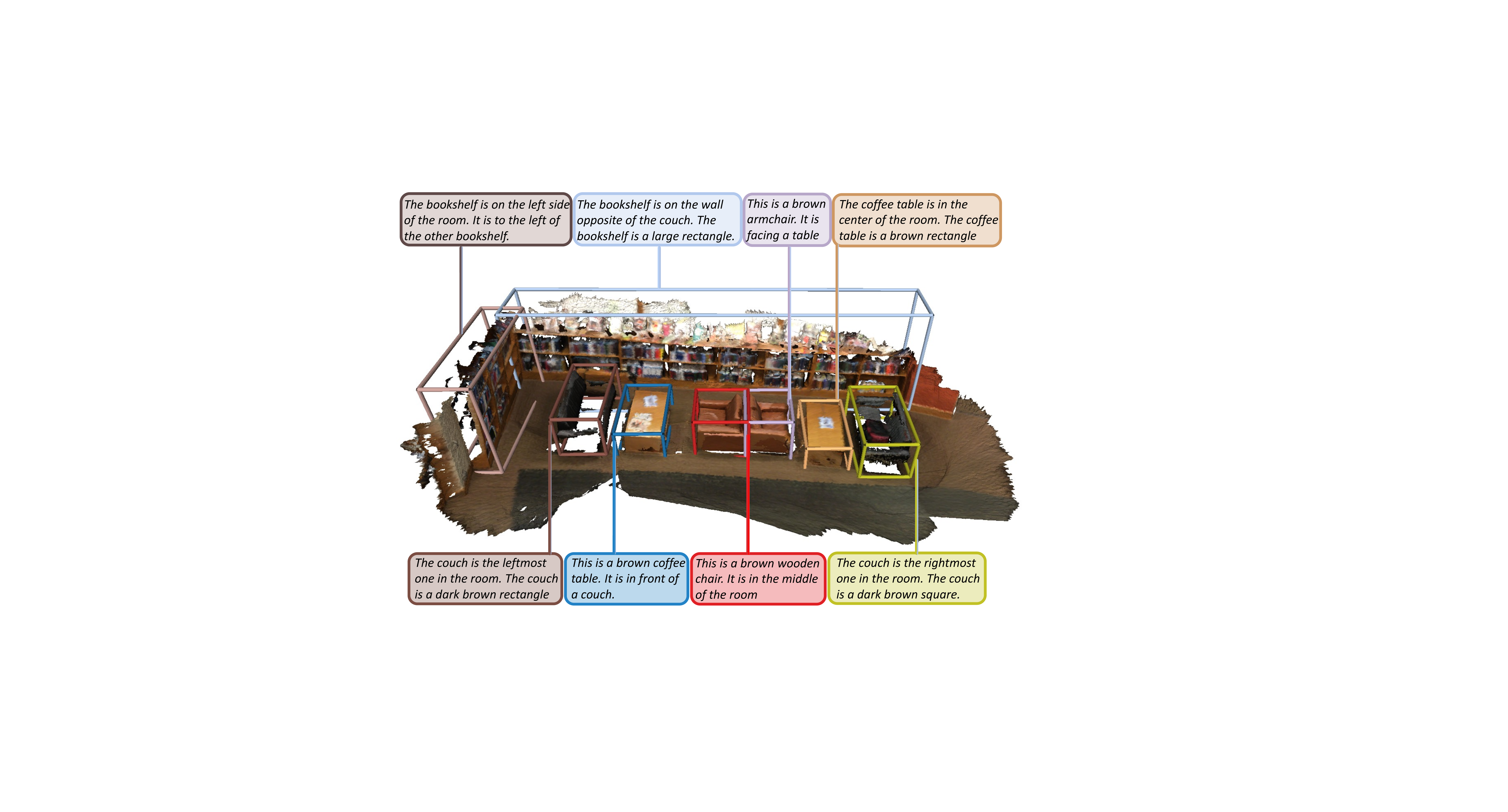}
        \caption{The illustration of dense captions for objects within a whole scene predicted by our method. We distinguish the captions for each object with different colors.}
        \label{figure:qualitative_scene}
    \end{minipage}%
    \hspace{12pt}
    \begin{minipage}[t]{0.45\linewidth}
        \centering
        \includegraphics[width=0.8\linewidth]{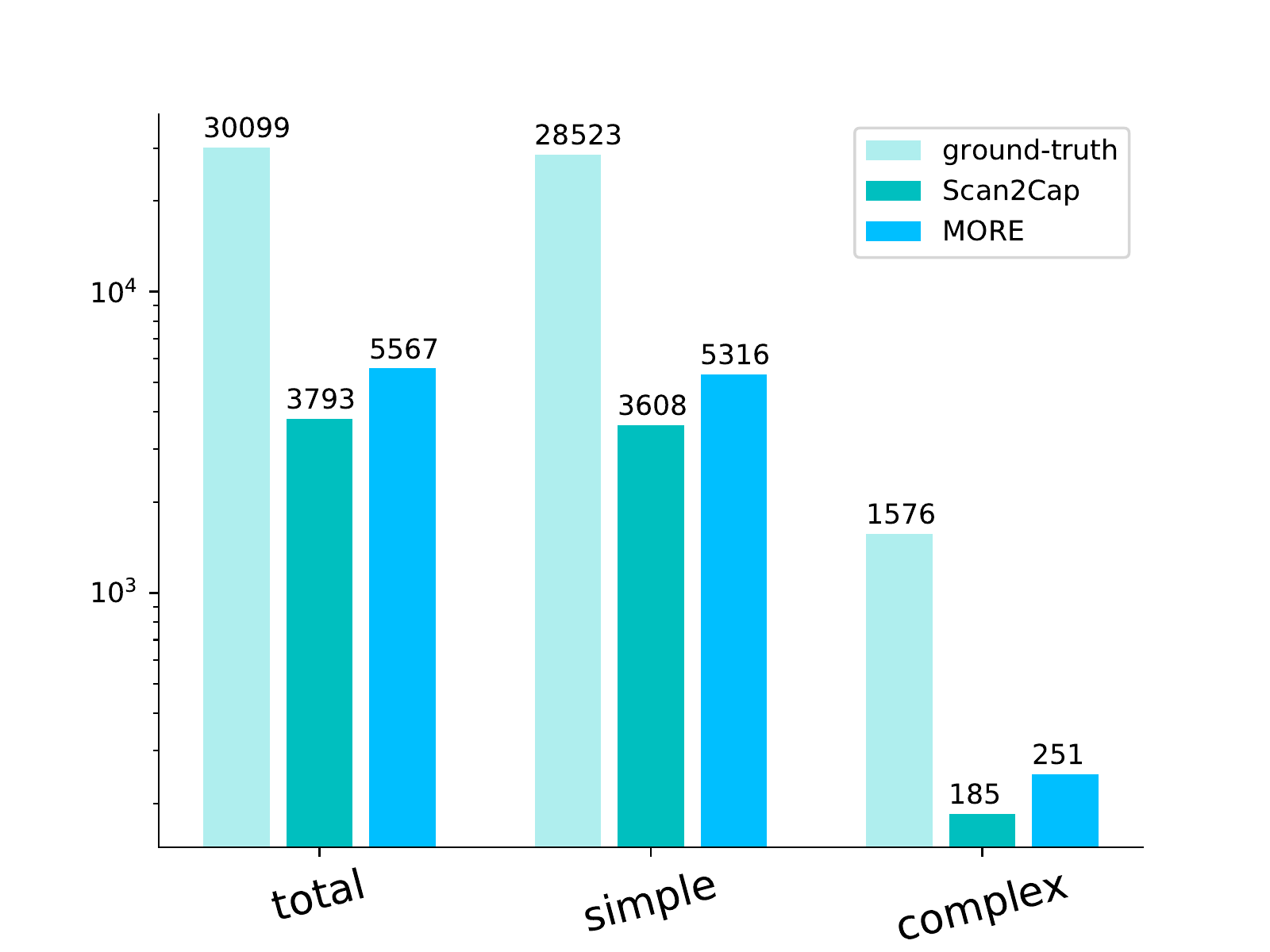}
        \caption{Statistics of relational words in captions generated by different methods. ``simple" and ``complex" roughly represents first- and multi-order relations, respectively, and ``total" is the sum of all relational words.}
        \label{figure:word_counts}
    \end{minipage}
\end{figure}

In order to quantitatively evaluate the distinctiveness of graph nodes, we calculate the MADGap, a metric introduced in~\cite{chen2020measuring} to evaluate over-smoothness of graph nodes.
The results are shown in the last column of Table.\ref{table:comprehensive}. 
As can be observed, when the number of stacked graph layers increases from 1 to 4, the corresponding MADGap value decreases, which indicates that the node features gradually become more similar. 
Although 1-layer SLGC can achieve higher MADGap value, its overall performances of caption generation are inferior to the 2-layer SLGC, which is mainly due to that the spatial relations are not sufficiently encoded into the nodes with only one graph layer.
When stacking more than 2 layers of SLGC, the MADGap and the overall captioning performances both decrease, demonstrating the over-smoothing is correlated with captioning quality degradation to some extent.  
Finally, combining 1-layer SLGC with our proposed OTAG, our method achieves much higher MADGap value than stacking any number of layers of SLGC, while also performs the best on caption generation.
This indicates that the object-centric graph construction in OTAG can preserve distinctive information of the node features, make the aggregated context of in the caption decoder more diverse, and finally boost the captioning performances. 

\noindent \textbf{Relational words statistics.} To give a more intuitive analysis on the advantages of our method, we inspect the performance improvement gained by our method in terms of relation capturing. Fig.\ref{figure:word_counts} compares the relational words in the sentences generated by Scan2Cap and our MORE, as well as in ground-truth annotations.
Specifically, we maintain a dictionary of all the relational words in the corpus. Then we classify these words into ``simple" and ``complex" according to whether multiple objects should be jointly considered to support predicting the relation. As can be seen, comparing to Scan2Cap, our MORE can capture more relations when describing an object, which demonstrates its effectiveness. The relational word dictionary, and the split of ``simple" and ``complex" words can be found in the supplementary materials. 

\subsection{Qualitative Results}
We further show the qualitative result in Fig.\ref{figure:qualitative_scene}. 
Note that to avoid clutter caused by low-quality object proposals, we directly use the ground-truth bounding box of each object.
We can observe that our method is able to capture diverse spatial relations among objects. 
For example, relations like ``on the left side of'', ``in front of'' and ``opposite of'' are all properly leveraged to describe objects. 
Besides, our method can also accurately describe multi-order relations, such as ``leftmost'' and ``rightmost'', which are used to describe the two couches at both ends of the scene. 
The results demonstrate that our MORE is effective in capturing and describing diverse spatial relations for multiple objects in 3D world.

\section{Conclusions}
In this paper, we improved 3D dense captioning by proposing a novel relation modeling method, named Multi-order RElation Mining Network (MORE). We progressively modeled spatial relations by encoding first-order and multi-order ones with our proposed Spatial Layout Graph Convolution (SLGC) and Object-centric Triplet Attention Graphs (OTAG), respectively. 
Extensive experimental results demonstrated the effectiveness and the advantage of MORE over the current state-of-the-art method.

\section*{Acknowledgements}
This work was supported in part by Shanghai Pujiang Program (No. 20PJ1401900) and Shanghai Science and Technology Program (No. 21JC1400600).

\clearpage

\bibliographystyle{splncs04}
\bibliography{reference}
\end{document}